%% file: paper.tex
\documentclass[11pt,a4paper]{article}
\usepackage[hyperref]{emnlp2020}
\usepackage{times}
\usepackage{latexsym}
\usepackage{xcolor}
\usepackage{url}
\usepackage{setspace}
\usepackage{xspace}
\usepackage{amsmath}
\usepackage{amsfonts}
\usepackage{booktabs}
\usepackage{subcaption}
\usepackage{graphicx}
\usepackage{mathtools}
\usepackage{soul}
\usepackage{bibunits}
\usepackage{mathabx}

\def\aclpaperid{445}
\aclfinalcopy

\usepackage{etoolbox}

\title{Beyond Instructional Videos:\\Probing for More Diverse Visual-Textual Grounding on YouTube}
\input{macros.tex}
\author{  \begin{tabular}{cc}
    Jack Hessel & \hspace{.31in} Zhenhai Zhu  \quad Bo Pang \quad Radu Soricut \\
    {\normalfont Allen Institute for AI} & {\normalfont Google Research} \\
    {\tt jackh@allenai.org} & {\tt \{zhenhai,bopang,rsoricut\}@google.com}
  \end{tabular}}
\date{}

\begin{document}
\maketitle

\begin{abstract}
  Pretraining from unlabelled web videos has quickly become the de-facto means of achieving high performance on many
  video understanding tasks. Features are learned via prediction of \emph{grounded} relationships between visual content and automatic speech recognition (ASR) tokens.
  However, prior pretraining work has been limited to only instructional videos; 
  \emph{a priori,} we expect this domain to be relatively ``easy:"
  speakers in instructional videos will often reference the literal objects/actions being depicted.
  We ask: can similar models be trained on more diverse video corpora?
  And, if so, what types of videos are ``grounded" and what types are not?
  We fit a representative pretraining model to the diverse YouTube8M dataset, and study its success and failure cases.
  We find that %
  visual-textual grounding is indeed possible across previously unexplored video categories, and that
  pretraining on a more diverse set results in representations that generalize to both non-instructional and instructional domains.
\end{abstract}

\maketitle

\begin{bibunit}[acl_natbib]

\section{Introduction}

Self-supervised pretraining approaches have recently been adapted to web videos \cite{sun2019contrastive,sun2019videobert,miech2019howto100m,miech2019end,zhuactbert,amrani2020noise}; the resulting models have achieved state-of-the-art performance on a wide range of video understanding tasks, e.g., dense caption generation, action localization, etc.

In general, the pretraining step requires a large, unlabelled corpus of web videos.
The training objective aligns visual content (i.e., video segments) with automatic speech recognition (ASR) tokens, and the resulting representations are fine-tuned for downstream tasks.
The assumption underlying this family of approaches is that, in the pretraining corpus, spoken words have \emph{some} consistent, grounded relationship with the temporally corresponding visual content.

However, in contrast to the highly diverse corpora utilized for text-based pretraining (Wikipedia, Common Crawl, etc.), pretraining for web videos (so far) has been limited to instructional videos.
This domain restriction is motivated by the commonly accepted notion that ``procedural knowledge tends to be inherently multimodal" \cite{malmaud2015s}.
We \emph{expect} that the semantic information in video frames and ASR tokens is readily correlated in instructional videos.
But corpus diversity brings significant benefits: in the text-only case, models can effectively represent diverse real-world entities \cite{roberts2020much} precisely because pretraining is not restricted to, e.g., only fictional stories \cite{zhu2015aligning}.

In search of more general representations, our main question is: \textbf{does video-ASR pretraining ``work" for more diverse pretraining corpora?}
Are certain categories of non-instructional videos ``groundable,'' thus enabling diverse representation learning?
Or are some types too difficult, only acting as training noise?
We conclude that:
1) grounding is indeed possible in a wide range of yet-to-be-computationally-exploited YouTube video categories, e.g., walk-throughs, vehicles, tech reviews, etc., with some harder than others;
2) transferable representations can be successfully learned by training on a more diverse set, which may provide more versatility.

\section{Related Work}
\label{sec:related_work}

ASR is known to be a useful signal source in various instructional video understanding tasks \cite{gupta2017visual,huang2017unsupervised,huang-buch-2018-finding-it,moriya2019grounding}, e.g., action detection/classification \cite{yu2014instructional,alayrac2017joint,chang2019d3tw,kuehne2019mining}, segmentation/captioning \cite{sener2015unsupervised}, and instruction alignment \cite{malmaud2015s,alayrac2016unsupervised}.
A number of multimodal instructional video datasets have been proposed \cite{wang2019youmakeup,tang2019coin,sanabria2018how2}.
A notable recent example of work addressing a non-instructional video corpus is \newcite{ignat2019identifying}, who analyze grounded-ness in lifestyle vlogs.
\newcite{fouhey2018lifestyle} highlight the difference between keyword search vs. implicitly mining action data of interest from a broader corpus (e.g., \newcite{bregler1997learning,gu2018ava}).

\mparagraph{Operational grounding} Our work builds upon prior operational notions of grounding: if an algorithm is able to consistently predict specific visual-textual relationships, then that relationship is said to be ``grounded" \cite{lu2008high,berg2010automatic,parikh2011interactively,hill2014learning,hessel2018quantifying}. %
\newcite{yanai2005image}, for example, 
examine an image+text corpus and
rank ``substrings of text by how well their occurrence can be predicted from visual features."
One shortcoming of any model-based operationalization of ``grounding'' is that only \emph{positive} instances of groundedness can be identified: if one model fails to ground something, perhaps a better model could have.

\section{Video-ASR pretraining + our model}
Recent work in designing pretraining objectives: 1) assumes that ASR tokens have, on average, some correspondence to temporally co-occurring video frames within the same video; and 2) ignores clips that lack ASR.
We consider a model that encapsulates both of these assumptions.\footnote{While more complicated models are possible, our goal is to conduct an error analysis of a simple, representitive model, not to necessarily achieve state-of-the-art results.}
The model is a slight simplification of \newcite{miech2019howto100m}, where a joint embedding for the visual content and ASR tokens is learned. While more sophisticated methods based on self-attention models have since been examined (e.g., \newcite{zhuactbert}), joint embedding models are still performant and offer greater interpretability, thus enabling our later error analyses.

\mparagraph{Model details} The similarity between \videoclip $i$ and \asrcaption $j$, $s_{i,j}$, is estimated by computing the cosine similarity between their corresponding embeddings in the joint space. Joint embedding models are parameterized using gated, multi-layer feedforward networks. The visual features we use as input are: frame-wise 2D Inception-v1 pretrained for object detection \cite{inception_v1,sun2017revisiting} and 3D CNN S3D-G features pretrained for action recognition \cite{xie2018rethinking,kay2017kinetics}. The language feature inputs are 300 dimensional vectors per word-type; these are fine-tuned during the training process. Max pooling is used for both token embeddings and for per-frame visual features to achieve a single visual and textual embedding for each clip.\footnote{When training on \yttrain the vocabulary size is 61K.}

During training, temporally corresponding (\videoclip, \asrcaption) pairs are sampled (``$\mathcal{P}$ositive" cases).
For each positive case, a set of mismatched ``$\mathbf{\mathcal{N}}$egative" cases is also sampled both from other videos and from the same video in equal proportion.
In contrast to \newcite{miech2019howto100m}, we control for clip length, and sample temporally \emph{fixed-length} \segments. In initial experiments with variable-length \segments, we found that our models were capable of ``cheating" the grounding task by aligning longer (and shorter, respectively) \videoclips with longer (and shorter) \asrcaptions, largely ignoring content. Thus, this simplifying choice makes our error analysis significantly more straightforward, and results in minimal performance change. We use 5 second \segments, but results are similar with 10 or 30 second windows (see \autoref{sec:window_size_stability}). To generate \segments, we initially randomly sample 256 per video before discarding ones that have no temporally-accompanying ASR. \Segments may overlap, though results are similar without overlaps (see \autoref{sec:window_overlap_stability}).
The following hinge loss is minimized for margin $\delta$:
\begin{equation}
\resizebox{1\hsize}{!}{ ${\displaystyle 
\sum_{i,j \in \mathcal{P} \text{, } \mathcal{N}}  \max(0, \delta + s_{i,j} - s_{i,i}) + \max(0, \delta + s_{j,i} - s_{i,i}) } $ }
\label{eq:loss_function}
\end{equation}

\noindent We trained with Adam \cite{kingma2014adam}, a learning rate of .001, and set $\delta=.1$, but didn't undertake significant hyperparameter optimization. We terminate training after 300K steps.

\mparagraph{CrossTask replication} To verify that our model simplifications didn't significantly hinder performance, we replicated key experiments from \newcite{miech2019howto100m}. In particular, we sought to gather the pretraining corpus they used, HowTo100M, which consists of 1.22M videos. Because of, e.g., users deleting videos, we were able to gather features for only 87\% of the original set, 1.06M videos.

\begin{figure}
    \centering
    \includegraphics[width=.95\linewidth]{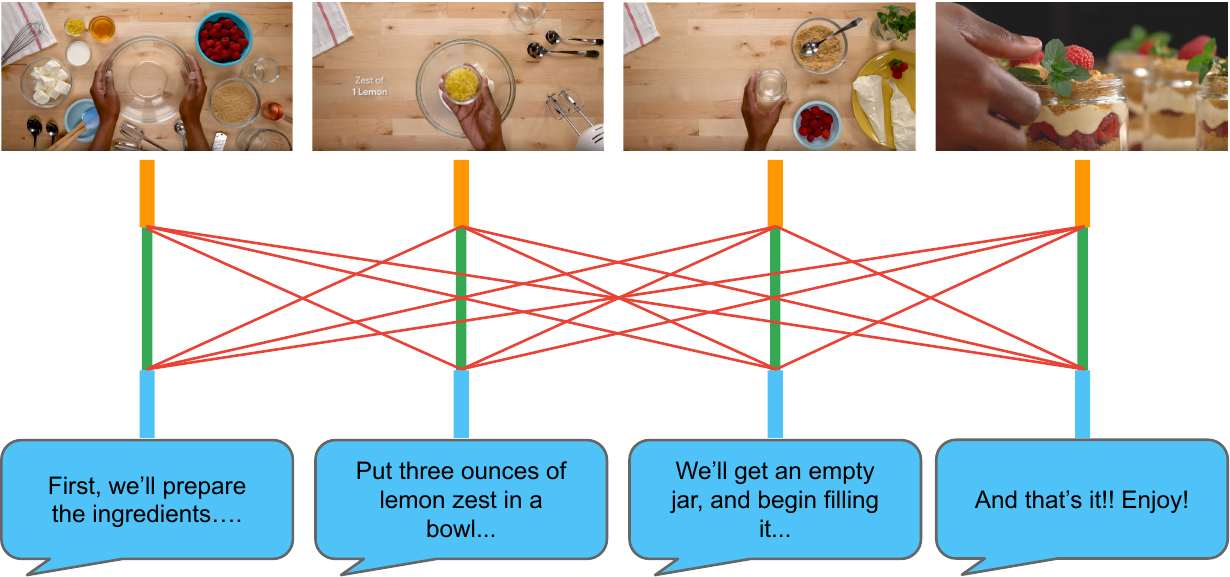}
    \caption{\Samevideoaucnospace metric: the model scores all possible links between \videoclips and \asrcaptions within a single video; the model is rewarded for assigning higher similarity to \textbf{{\color{darkgreen}temporally-aligned \segments}} versus \textbf{{\color{red}mismatched ones}}. }
    \label{fig:vidlevelauc}
\end{figure}

We verify the performance of our model using the CrossTask localization task \cite{zhukov2019cross}. While we defer details to the original paper, the goal of CrossTask is to temporally localize a set of procedural steps for a task in an unlabelled/unsegmented video depicting that task. An algorithm's performance is evaluated with a recall metric (higher is better). We follow the evaluation procedure given in \newcite{miech2019howto100m}, except instead of embedding each frame individually, we embed a sliding 5-second window of video clips.

Our simplified model trained on less data performs comparably to \newcite{miech2019howto100m}'s. We achieve 32.6 recall, while they report 33.6 recall; for reference, a supervised upper-bound without pre-training achieves 31.6 recall (full results and more details are in \autoref{sec:sec_with_full_crosstask_results}). 

\mparagraph{Measuring visual-textual alignment} 
Viewed through the lens of link prediction between truly co-occuring (clip, ASR) pairs, Eq.~\ref{eq:loss_function} can be seen as a differentiable approximation of \auc~\cite{rendle2012bpr}.
Thus, we propose to operationally measure the groundedness using 
\emph{\samevideo} \auc:
a single score is assigned to each video, rewarding the model if it is able to successfully align temporal pairs within the same video (and penalizing it if not).
Fig.~\ref{fig:vidlevelauc} presents a visualization of this method. One notable advantage of \auc versus other link prediction metrics is that it is insensitive to the label distribution: shorter videos are not systematically assigned higher scores simply because there are fewer incorrect links.

\section{A More Diverse Corpus} \label{sec:youtube8m_asr}

\mparagraph{\yttrain}
YouTube8M \cite{abu2016youtube} is a dataset of 6.1M YouTube videos,\footnote{v3 of the dataset is smaller than v1/v2, due to videos becoming unavailable over time and other refinements.} where each video is labeled across 3K categories, ranging from ``cooking'' to ``games'' to ``nature." 
It is among the largest and most diverse publicly available dataset of YouTube videos.
Due to user deletions and videos without detected spoken words, we are able to collect ASR via the YouTube API
for 1.4M (29\%) videos; we further filtered to 817K videos tagged with English ASR.\footnote{We expect that non-English videos will similarly be an excellent source of visual-textual grounding training data, particularly for under-resourced languages. We focused on English to simplify our error analyses. But in future work, we expect to not impose such a limitation.}
There is an extremely wide variance of ASR availability per category, e.g., 74\% of ``BBC" videos (a category that generally contains news videos by the broadcaster) have ASR, whereas almost no ``Military band" videos do (Table \ref{tab:asr_per_category}). While the percentage of ASR-available videos is higher in many instructional video categories, e.g., ``cooking" at 31\%, ``cosmetics" at 44\%, etc., many non-instructional categories on YouTube have ASR available (e.g., ``silver" at 65\%; mostly videos about coins).
Maintaining the train / validation split of the original data release yields 639K training  videos (henceforth referred to as \yttrain) and 167K validation-set videos.

\begin{table}[t]
    \hfill
    \begin{subtable}[t]{0.45\linewidth}
        \centering
        {\scriptsize
        \input{tables/asr_category_top.tex}
        }
        \caption{Most ASR}
        \label{tab:most_asr_categories}
    \end{subtable} 
    \hfill %
    \begin{subtable}[t]{0.45\linewidth}
        \centering
        {\scriptsize
        \input{tables/asr_category_bottom.tex}
        }
        \caption{Least ASR}
        \label{tab:least_asr_categories}
    \end{subtable}
    \hfill

    \caption{Categories of YouTube8M with the highest and lowest availability of English ASR (minimum 1K videos); corpus mean = 17\%.}
    \label{tab:asr_per_category}
\end{table}

\mparagraphnp{Human annotation of ``Is-it-instructional''}
While a qualitative examination of YouTube8M reveals clear topical and stylistic diversity compared to domain-restricted corpora, we quantitatively verify that YouTube8M does not consist of mostly instructional videos. 

We sample 6.8K videos with English ASR from the validation set for human labeling.
Each video is shown to three paid annotators, who each provide a Yes/No answer to the question:
``Does this video focus on real-world human actions accompanied by procedural language that explains what is happening on screen in reasonable detail?'' %
Note that our definition of ``instructional'' intends to include the usual ``how-to'' videos, but also attempts to capture a more general notion of ``instructional-ness''.
For instance, an un-boxing video where parts of a product are {\em taken out and assembled} along with corresponding narration should receive ``Yes'', whereas a video showing only a product from different angles should receive ``No'', due to a lack of narrated human actions.  %

After a pilot study with a few iterations over the guidelines and examples, the annotators reach high agreement: in 96\% of cases, all three judges are unanimous.
From these annotations, we estimate that around 74\% of the videos in the \yttrain corpus are \textbf{not} instructional, even for the generalized notion of ``instructional-ness.''
For reference, \newcite{miech2019howto100m} conduct an analysis of 100 videos from HowTo100M (constructed with the intention to focus on how-to videos) and estimate that 71\% are instructional. %

The annotated {\em i3-video} corpus (\underline{\textbf{i}}s-\underline{\textbf{i}}t-\underline{\textbf{i}}nstructional-video) is available for download.\footnote{ \url{https://github.com/google-research-datasets/i3-video}} One potential use-case: consider an automated tool designed exclusively for use on instructional videos. A classifier trained on our labelled corpus could be used to determine if applying the automated tool is appropriate or not for an unlabelled input video.

\begin{figure}
    \centering
    \includegraphics[width=.9\linewidth]{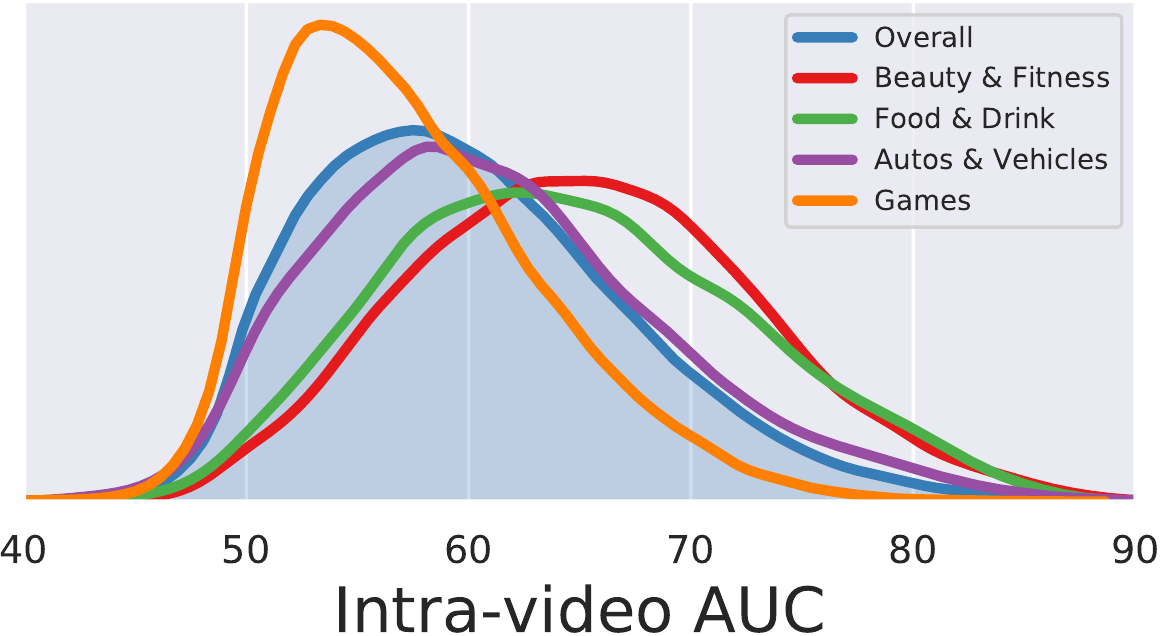}
    \caption{Distribution of \samevideoauc scores, grouped by meta category, compared to the overall distribution.}
    \label{fig:kde_meta}
\end{figure}

\mparagraphnp{Which categories are easiest/hardest?}
We train our model on \yttrain, and compute \samevideoauc for each of the 178K validation videos.
First, we average all videos labeled with a particular category to produce a per-category \auc score.
The performance in a vast majority of categories is above the 50 \auc (random) baseline, and ranges from 51 (``Mixtape") to 76 (``Muffin"). To make sure that the model is not succeeding simply because a category happened to be frequent in the dataset, we note the correlation between category \auc and category frequency is 
essentially zero ($\rho = .02, p > .58$).
This suggests that at least \emph{some} aspect of most categories of videos can be visual-textually grounded.

We next coarsely aggregate the YouTube8M categories into meta-categories, e.g., ``Food and Drink."\footnote{These meta-categories are called ``verticals," and are released with YouTube8M.}
The \auc distribution of 4 popular meta categories relative to the overall \auc distribution is given in Fig.~\ref{fig:kde_meta}.
In general, the grounding succeeds most readily on makeup/hair videos (e.g., ``Eye liner" $\aucmath = 74$, ``Updo" $\aucmath = 68$, etc.) and cooking videos (e.g., ``Vegetarian cuisine" $\aucmath=71$), domains that have been previously used in video grounding work.
Besides these already-studied domains, other high-scoring category types emerge (Table~\ref{tab:new_domains}).
Conversely, some categories are more difficult for the model, e.g., video game categories like ``RuneScape" $\aucmath=54$ and ``First-person shooter" $\aucmath=55$; speakers in these videos often reference diverse topics unrelated to the game itself.
Non-video-game categories can also be difficult, e.g., ``Unidentified flying object" $\aucmath=56$, 
``Dashcam" $\aucmath=54$.

\begin{table}
    \centering
    \input{tables/new_domains.tex}
    \caption{Domains with high operational groundability.}
    \label{tab:new_domains}
\end{table}

We next ask: \emph{are instructional videos indeed easier to operationally ground?} %
While human judgements of instructional-ness and \samevideoauc are positively correlated $\rho = .20$ ($p \ll 0$), the low magnitude of this correlation provides additional empirical confirmation that other types of videos are also promising.

\mparagraph{Within-category observations}
To this point, we have identified broad categories of YouTube videos that are more groundable than others.
However, it is not yet clear \emph{why}, e.g., the algorithm gets 64 \auc on ``Action Figure," or 55 \auc on ``Call of Duty" (a first-person shooter game).
We now define a \emph{segment-level} \auc metric, analogous to the \samevideoauc metric previously defined: it quantifies how readily \emph{individual \asrcaptions} are temporally localized by the model within the same video (see \newcite{menon2011link} for a description of different \auc variants).

\begin{figure}
\centering
    \begin{subfigure}{.95\linewidth}
    \centering
     \includegraphics[width=0.48\linewidth]{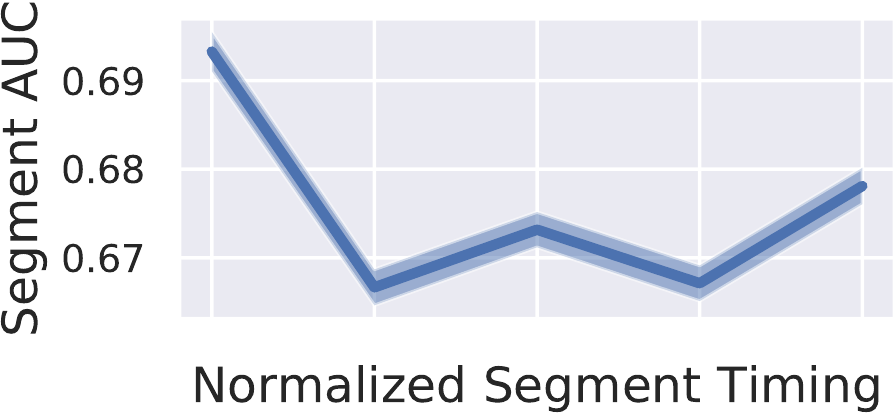}
     \includegraphics[width=0.48\linewidth]{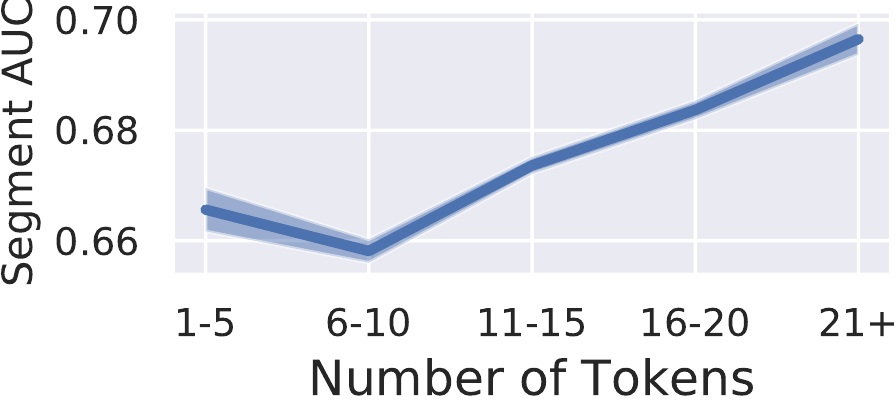}
    \caption{Action Figure (64 category \auc)}
    \label{fig:action_figure_length_timing}
    \end{subfigure}
    
    \begin{subfigure}{.95\linewidth}
    \centering
     \includegraphics[width=0.48\linewidth]{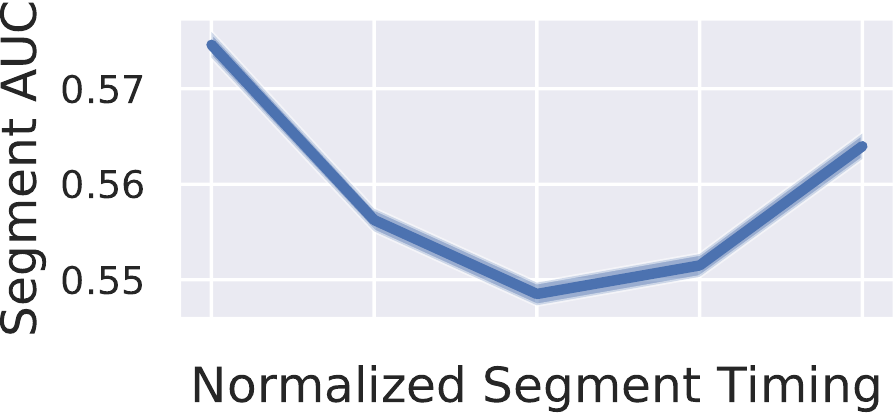}
     \includegraphics[width=0.48\linewidth]{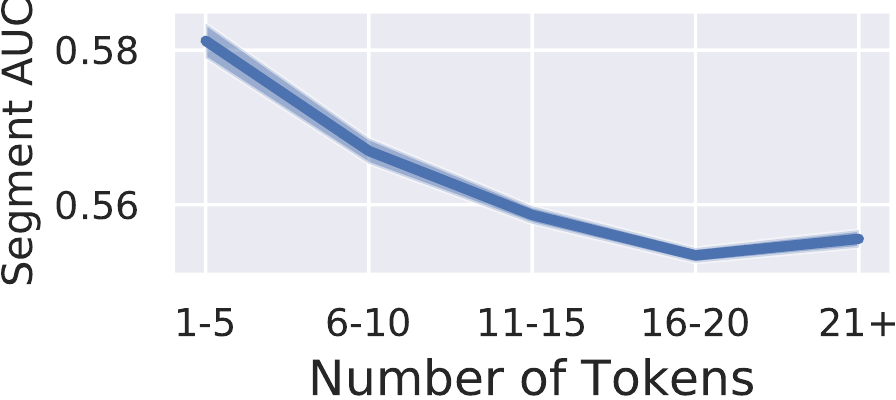}
    \caption{Call of Duty (55 category \auc)}
    \label{fig:call_of_duty_length_timing}
    \end{subfigure}
    \caption{Correlation between per-segment \auc scores and \segment timing within video (left column) and the number of tokens in a \segment (right column)}
    \label{fig:length_and_timing}
\end{figure}

\label{sec:sec_with_action_figure}

Before examining the relationship between content-based features and segment-level \auc, contextual factors must be considered.
Fig.~\ref{fig:length_and_timing} illustrates clear relationships 1) between \asrcaption placement within a video and segment \auc (segments at the very beginning and very end of videos tend to be easier); and 2) between the number of tokens in an \asrcaption and segment \auc. For ``Action Figure", ASR segments with more words are easier (this is the case with most categories), but for ``Call of Duty", the opposite is true.

After controlling for contextual variables, we train OLS regression models to predict segment \auc from lexical unigram features, while controlling for timing/length features.
Lexical features add predictive capacity ($p \ll .01$, F-test).
While we find some patterns predictive of segment \auc for both categories, e.g., intro/outro-language (e.g., ``hey", ``welcome", ``peace"), we also observe topical patterns, e.g., several unigrams associated with specific action figure body parts (``knee", ``shoulder", ``joint", etc.) are positively associated with segment \auc.

\section{Implications for Pretraining}
While we've thusfar shown that self-grounding is possible for a diverse set of domains, do we \emph{gain} anything by training on a more diverse corpus?
Or do difficult-to-ground videos introduce noise and degrade representations for downstream tasks?

We compare two versions of our model: one with parameters learned from training on the diverse \yttrain corpus (\diversem), and one with parameters learned from a domain-specific corpus of 1M instructional videos (\instrm).

First, we evaluate each model's capacity to localize instructional steps on the CrossTask \cite{zhukov2019cross} dataset. \diversem performs admirably, even with significant domain mismatch and fewer pretraining videos: recall drops by only 15\% ($32.6 \rightarrow 27.4$) when swapping from \instrm to \diversem.

We next evaluate each model's performance on the same-video clip alignment task over a diverse set of videos: the sample of 6.8K human-annotated videos from the YouTube8M validation set.
In terms of \samevideoaucnospace, \diversem outperforms \instrm on 59\% of videos. %
If we split the data across the ``Is-it-instructional'' human judgements and compare the two models in each subset,
\instrm ``wins'' in 57\% of the instructional videos, whereas \diversem ``wins" in 65\% of non-instructional cases.

\mparagraphnp{In short:} both models achieve reasonable performance under instructional vs. non-instructional train/test domain mismatch. Taken together, this is a promising result for future pretraining work with more diverse corpora: at least for these evaluations, good performance on an instructional video grounding task is still possible under domain shift.
And while the comparison of \samevideoaucnospace is not necessarily definitive, it suggests that diverse corpora may provide more versatility, and we look forward to exploring this further in future work.

\section{Conclusion}

Peeking through the lens of a joint embedding model,
we probe into learning visual-textual grounding over a more diverse corpus of YouTube videos vs. prior work.  We find that learning visual-textual grounding is possible across many yet-to-be-explored categories of YouTube videos, and that it's possible to learn generalizable representations from a more diverse video set.

\mparagraph{Acknowledgements}
In addition to the anonymous reviewers,
the authors would like to thank 
Soravit Changpinyo,
Sebastian Goodman,
Bryan Seybold,
Chen Sun and
Ashish Thapliyal 
for insightful discussions
and implementation help. JH completed this work during an internship at Google.

\putbib[refs]
\end{bibunit}

\clearpage

\begin{bibunit}[acl_natbib]
\appendix

\section{Additional Model Details}

We adapt \newcite{miech2019howto100m}'s joint embedding model that pre-trains by aligning ASR tokens with corresponding video frames. The main difference between our implementation and theirs is how we generated (ASR, caption) pairs. While we considered generating clips according to their methodology, we ran into two problems. First, in early experiments, we found that the interpretability our error analysis was significantly impacted by varying clip length. For example: we were worried that it might not be consistent to compare the model's ability to temporally ground a 1s clip vs. a 15 second clip. There was also high correlation between caption length and temporal clip duration, which further complicated interpretation. Sampling clips of uniform duration solved these problems.

Second, \newcite{miech2019howto100m}'s temporal segmentation was generated by relying on the \emph{scrolling timing} of the ASR tokens on the YouTube, i.e., the time that YouTube decides to generate a linebreak, removing a line of caption from the screen. Via manual inspection, we found that scrolling time was temporally unreliable, e.g., the time in which ASR captions scroll on YouTube often differs significantly from when particular words were said. Instead, we sample 256 candidate 5 second \segments uniformly at random from the video, and then discard \segments that have no corresponding ASR.

\mparagraph{Additional visual feature details} For 2D features, we sample frames at 1FPS from all of the videos in our corpus, resize frames to be 256 by 256, and pass them through Inception-v1 \cite{inception_v1}
pretrained on JFT \cite{sun2017revisiting}. For 3D convolutional networks, we follow a similar procedure to \cite{sun2019videobert}, sample frames at 30FPS, aggregate frames into one second non-overlapping clips of 1 second each, and run an S3D-G \cite{xie2018rethinking} network that is pretrained on the Kinetics action recognition dataset \cite{kay2017kinetics}.
Both 2D and 3D features are L2 normalized.  The result of this process is a 2524-D feature vector for each second of video in our corpus.

\section{Comparison to HowTo100M}
\label{sec:sec_with_full_crosstask_results}

The full per-task recall comparisons are given in Table~\ref{tab:crosstask_replication}. Our results, like those of \newcite{miech2019howto100m}, use \newcite{zhukov2019cross}'s dynamic programming postprocessing method. We found that it usually resulted in a small performance increase.

Our simplified model performs only slightly worse (3\%) than \newcite{miech2019howto100m}'s. While we argue that our model is certainly still representative, there are several reasons why this gap might exist. For example,
there may be a regularizing effect when the model is allowed to view clips of varying length. Furthermore, our feature set was different; we used different (but comparable) base neural networks for feature extraction. Also, our model is trained on less data due to authors deleting their videos. Finally --- we didn't tune the training hyperparameters for our model/implementation, e.g., hinge size, learning rate, batch size, etc.
\input{tables/crosstask_results.tex}

\section{Stability of results to checkpoint}
\label{sec:checkpoint_stability}
\begin{figure}
\centering
    \begin{subfigure}{.95\linewidth}
    \centering
     \includegraphics[width=0.48\linewidth]{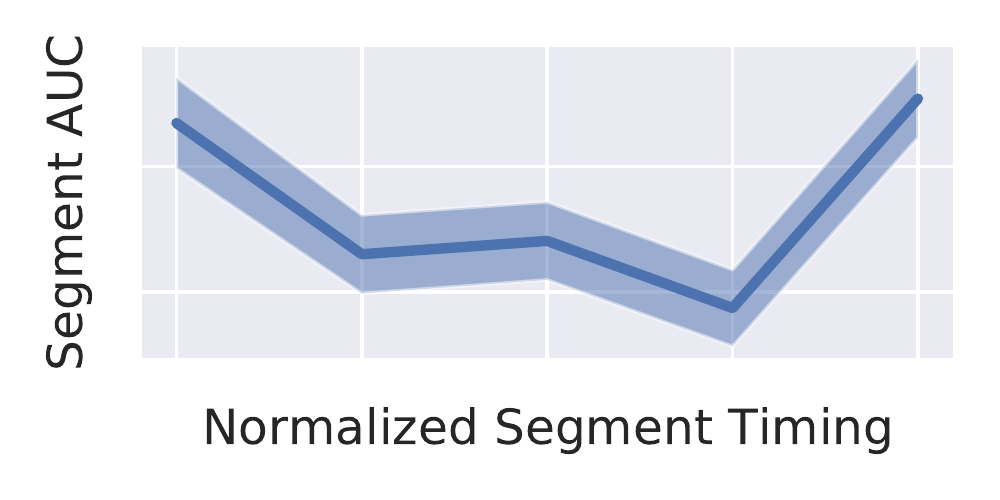}
     \includegraphics[width=0.48\linewidth]{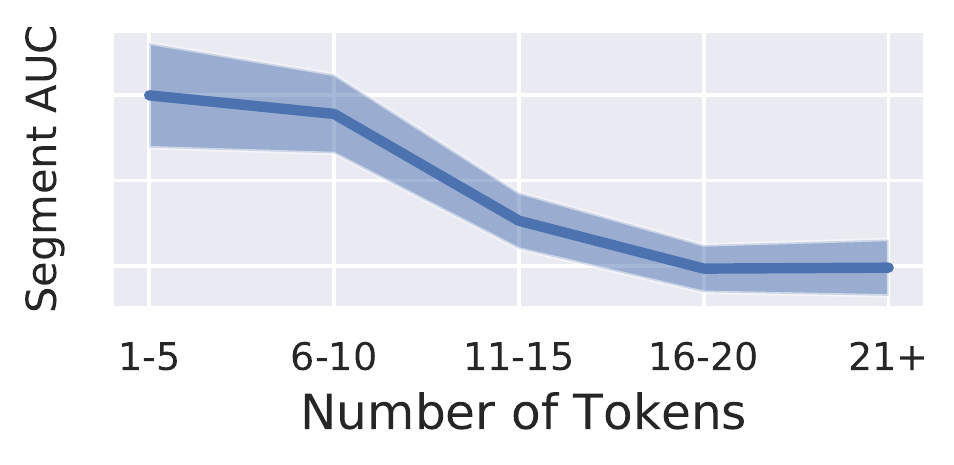}
    \caption{Action Figure (64 category \auc)}
    \label{fig:action_figure_length_timing_checkpoint}
    \end{subfigure}
    
    \begin{subfigure}{.95\linewidth}
    \centering
     \includegraphics[width=0.48\linewidth]{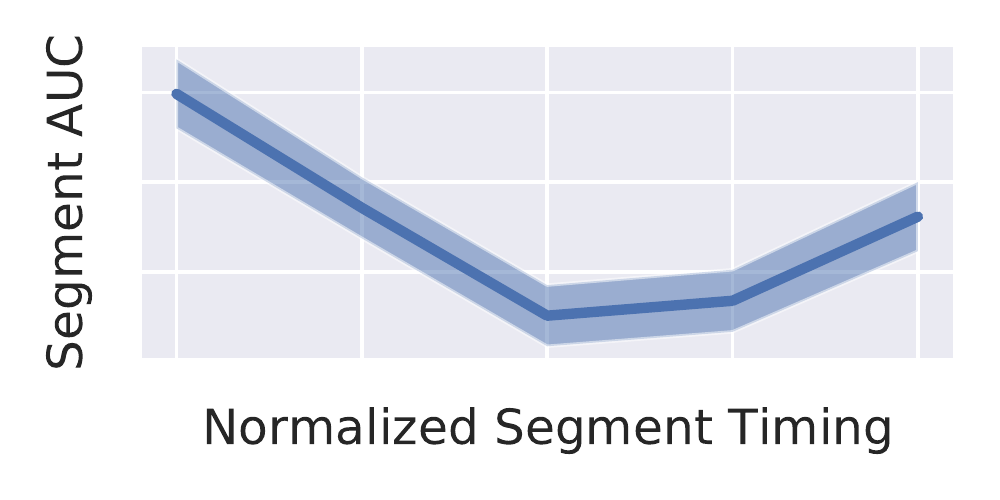}
     \includegraphics[width=0.48\linewidth]{figures/supplementary_figures/callofdutylength_by_auc.pdf}
    \caption{Call of Duty (54 category \auc)}
    \label{fig:call_of_duty_length_timing_checkpoint}
    \end{subfigure}
    \caption{Correlation between per-segment \auc scores and \segment timing within video (left column) and the number of tokens in a \segment (right column) for the 140K checkpoint (compare to the 300K checkpoint in the main paper).}
    \label{fig:length_and_timing_checkpoint}
\end{figure}

To ensure the results related to \samevideoaucnospace were insensitive to the particular choice of model checkpoint, we re-did the experiments in \S\ref{sec:sec_with_action_figure} using a version of our model checkpointed at 140K iterations
vs. the 300K presented in the main paper; these experiments were conducted over 21K
dev videos instead of the full 167K dev videos presented in the main paper. Figures+tables in that section were consistent with the presented results, and the qualitative observations about the ``Action Figure" category held (see Figure~\ref{fig:length_and_timing_checkpoint} for replicated figures).

\section{Stability of results to overlapping windows}
\label{sec:window_overlap_stability}
When we generate our windows at training and testing time to compute \samevideoaucnospace, given that we sample 256 candidates per video (and then filter out clips without associated temporal ASR),  the windows frequently overlap. We ran additional experiments to ensure that our results held when we sampled non-overlapping clips at testing time.

\begin{figure}
\centering
\includegraphics[width=0.9\linewidth]{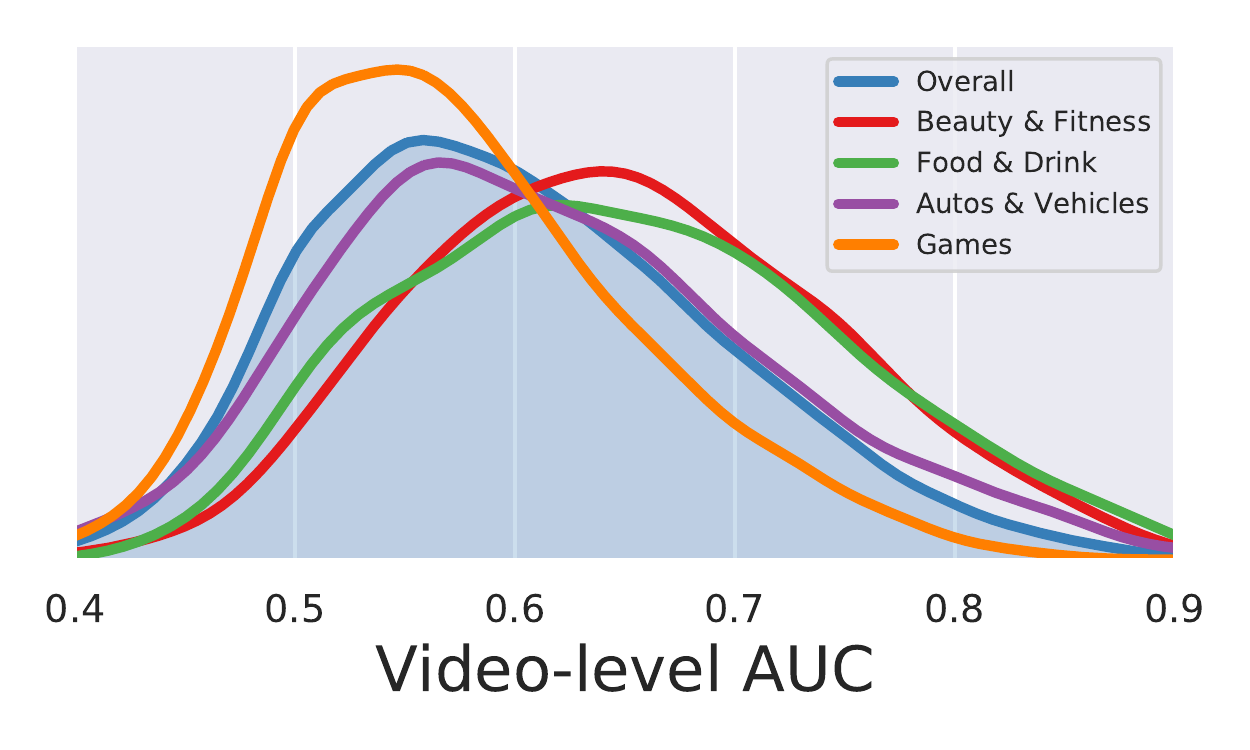}
\caption{Meta-category AUC with window size of 5, but test-time windows are sampled without overlap.}
\label{fig:meta_nooverlap}
\end{figure}

We computed an alternate version of the \samevideoauc results using a 140K training iteration checkpoint. Instead of sampling 256 segments at testing time, we only (randomly) sample up to 10 segments, but force them to be non-overlapping. There are some videos that are discarded in this process. For 1/500 videos (or so) we cannot sample non-overlapping segments. However, among the majority of videos for which the sampling is successful, the Spearman correlation with the category-level results allowing for overlap is $\rho=.98$. Figure~\ref{fig:meta_nooverlap} reproduces the meta-category plot from the main paper, but with test-time segments sampled without overlap.

\section{Additional Window Sizes}
\label{sec:window_size_stability}
The results presented in the main paper use a temporal window size of five seconds. We were curious as to the stability of our observations with respect to the choice of this window size. While changing the window size, to an extent, changes the nature of the task, we ran with window size 10s and window size 30s to measure the stability of the results.

\begin{figure}
\centering
    \begin{subfigure}{.95\linewidth}
    \centering
     \includegraphics[width=0.9\linewidth]{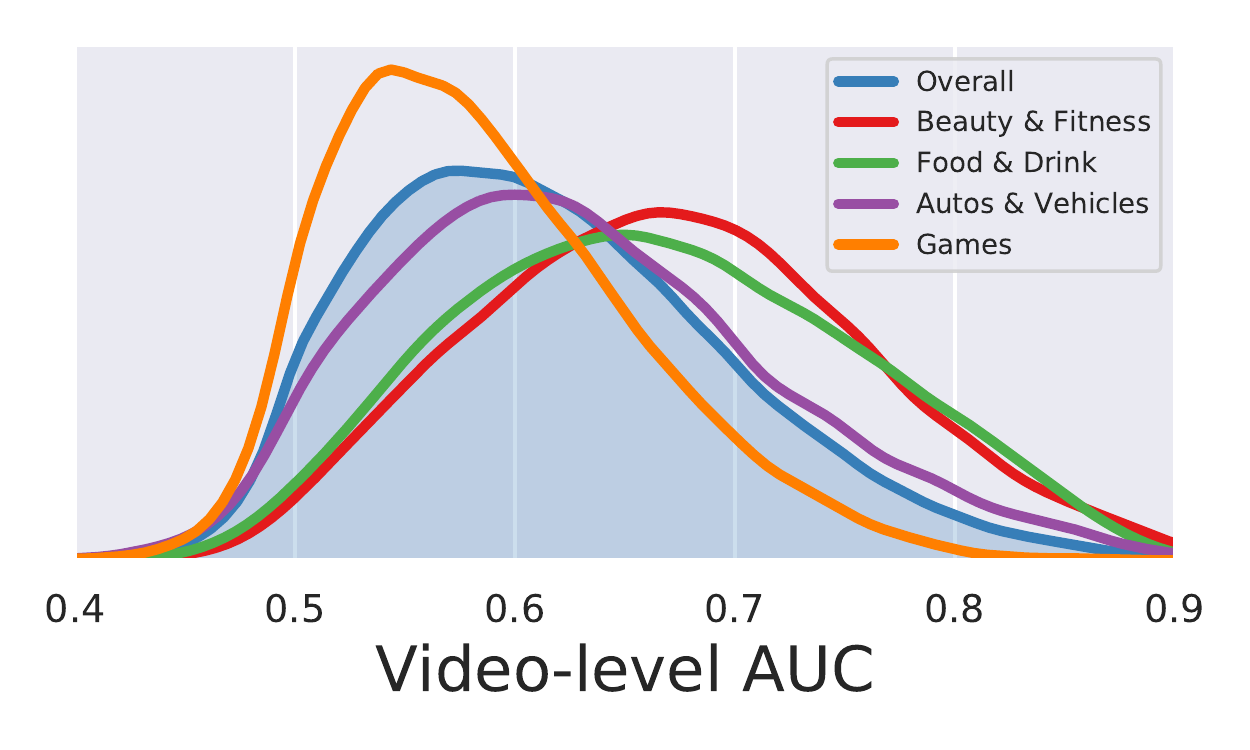}
    \caption{Window size = 10}
    \label{fig:meta_window_size_10}
    \end{subfigure}
    
    \begin{subfigure}{.95\linewidth}
    \centering
     \includegraphics[width=0.9\linewidth]{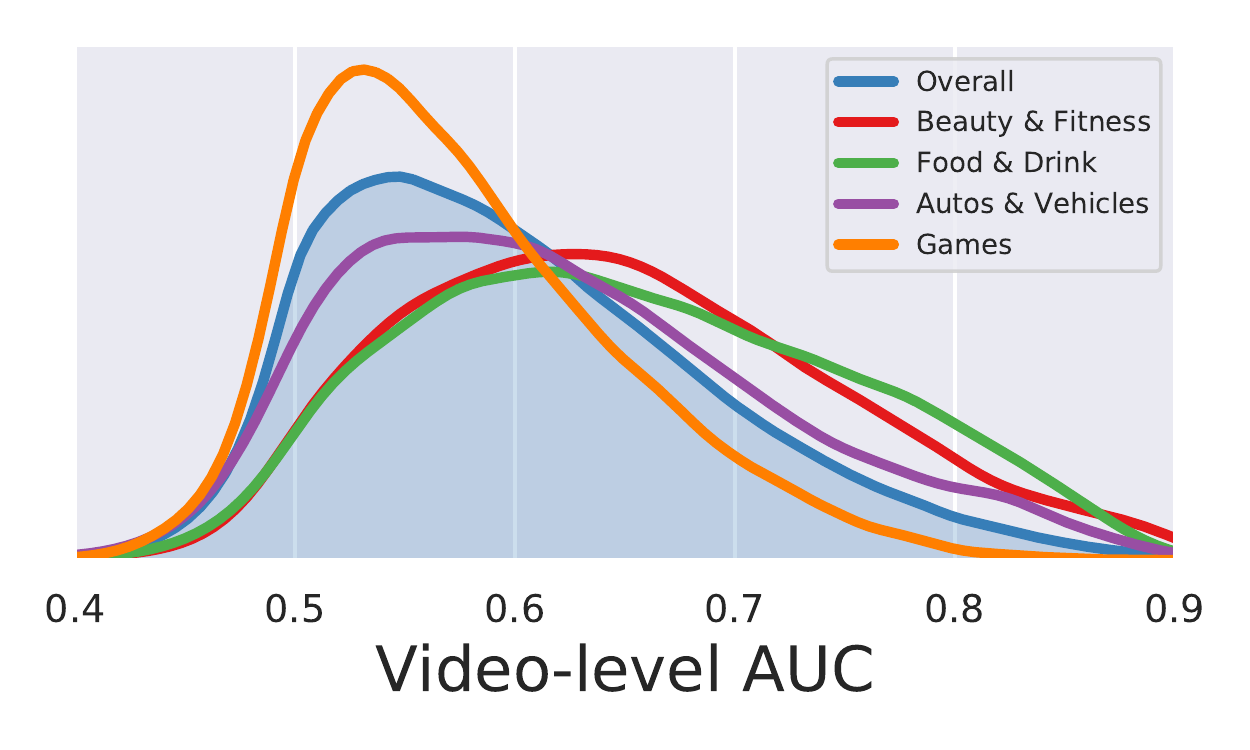}
    \caption{Window size = 30}
    \label{fig:meta_window_size_30}
    \end{subfigure}
    \caption{Meta-category AUC using models trained with alternate window sizes of $t=10,30$ seconds. The main paper results are with $t=5$ second windows.}
    \label{fig:length_and_timing_window_sizes}
\end{figure}
\mparagraph{Category-level \auc} We computed the \samevideoauc values for the models trained with alternate window sizes. Largely, while the individual \auc values may change by a point or two, the relative pattern stayed the same. The per-category Spearman correlation with the results for the $t=5$ windows was high: $\rho=.99$ for $t=10$ and $\rho=.96$ for $t=30$ (both correlations are true with $p \ll .001$). It should be noted that there are marginal differences in the development set videos used for the experiments at $t=5,10,30s$: because experiments were run at different times, due to video deletion, 2\% fewer videos were available for later experiments. Figure~\ref{fig:length_and_timing_window_sizes} recreates the meta-category plot from the main paper with the different window sizes; Tables~\ref{tab:new_domains_10} and ~\ref{tab:new_domains_30} report the same category-level \auc values as the table in the main paper. Overall, the results are very similar.

\mparagraph{Transfer learning} In the main paper, we explored the differences between models trained on \yttrain vs. HowTo100M. Here, we conducted the same YouTube8M self-grounding experiments as described in the main paper (varying the training set) with $w=10$ instead of $w=5$. The results are very similar.
In terms of \samevideoaucnospace, \diversem (trained on \yttrain) outperforms \instrm (trained on HowTo100M) on 58\% of videos (compared to 59\% for $w=5$ in the main paper).
If we split the data across the ``Is-it-instructional'' human judgements and compare the two models in each subset,
\instrm ``wins'' in 57\% of the instructional videos (compared to 57\% for $w=5$ in the main paper), whereas \diversem ``wins" in 63\% of non-instructional cases (compared to 65\% for $w=5$ in the main paper).

\begin{table}
    \centering
    \input{tables/supplementary_tables/10_categories.tex}
    \caption{High groundability categories with $t=10$ seconds.}
    \label{tab:new_domains_10}
\end{table}

\section{Additional Reproducability Information}
The models were trained and evaluated on a mix of TPU and GPU clusters. Depending on the particular hardware, the training process takes roughly 1 or 2 days to get 300K training iterations. While we make no particular claim about parameter/time efficiency, the number of parameters of our models are similar to HowTo100M's, i.e., roughly 50M. The runtime of our model is relatively fast --- on GPUs, batches of hundreds of videos can be processed in seconds. For hyperparameters not specifically described, we mirror the choices made in the HowTo100M public repo.\footnote{\url{https://github.com/antoine77340/howto100m}} For evaluating CrossTask, we wrote our own recall computing code that mirrors the setup publicly released by the authors;\footnote{\url{https://github.com/DmZhukov/CrossTask}} this includes several ad-hoc decisions, e.g., computing the floor/ceiling of temporal annotations in CrossTask to form the window, and only counting a "hit" when the predicted time is strictly \emph{less} than the ceiling (rather than less than or equal).
\begin{table}
    \centering
    \input{tables/supplementary_tables/30_categories.tex}
    \caption{High groundability categories with $t=30$ seconds.}
    \label{tab:new_domains_30}
\end{table}

\putbib[refs]

\end{bibunit}

\end{document}

%% file: macros.tex
\newcommand{\mparagraph}[1]{\noindent\textbf{{#1}}.}
\newcommand{\mparagraphnp}[1]{\noindent\textbf{{#1}}}

\newcommand{\aucmath}{{\scriptstyle \sf{AUC}}}
\newcommand{\auc}{$\aucmath$\xspace}

\newcommand{\videoclip}{clip\xspace}
\newcommand{\videoclips}{clips\xspace}

\newcommand{\asrcaption}{ASR caption\xspace}
\newcommand{\asrcaptions}{ASR captions\xspace}

\newcommand{\segment}{segment\xspace}
\newcommand{\segments}{segments\xspace}

\newcommand{\Segments}{Segments\xspace}

\newcommand{\samevideo}{intra-video\xspace}

\newcommand{\samevideoauc}{\samevideo \auc\xspace}
\newcommand{\samevideoaucnospace}{\samevideo \auc}

\newcommand{\Samevideoaucnospace}{Intra-video \auc}
\newcommand{\yttrain}{YouTube-600K\xspace}

\newcommand{\diversem}{$M_{\scriptstyle\sf{Diverse}}$\xspace}
\newcommand{\instrm}{$M_{\scriptstyle\sf{Instructional}}$\xspace}

\definecolor{darkgreen}{HTML}{256B38}

\newif\iffinal

\finalfalse

\iffinal
  \newcommand{\jack}[1]{}
  \newcommand{\bo}[1]{}
\else
  \newcommand{\jack}[1]{\textcolor{red!55!blue}{\textbf{Jack: #1}}}
  \newcommand{\bo}[1]{\textcolor{green!55!blue}{\textbf{Bo: #1}}}
\fi

%% file: tables/asr_category_top.tex
\begin{tabular}{l@{\hspace{.2cm}}c}
    \toprule
    Category & \% videos  \\
    \midrule
BBC & 74.1\% \\
President (USA) & 71.7\% \\
Hair conditioner & 71.1\% \\
Madden NFL & 69.3\% \\
Wig & 67.9\% \\
Magic (card game) & 67.9\% \\
Booster pack & 67.4\% \\
Raw foodism & 66.5\% \\
NBA 2K14 & 65.2\% \\
Silver & 65.2\% \\
    \bottomrule
\end{tabular}

%% file: tables/asr_category_bottom.tex
\begin{tabular}{l@{\hspace{.2cm}}c}
    \toprule
Category & \% videos  \\
    \midrule
Pachinko & 0.4\% \\
Jumbotron & 0.4\% \\
Chipmunk & 0.3\% \\
Taiko no Tatsujin & 0.2\% \\
Yo-kai Watch & 0.1\% \\
Zee Bangla & 0.1\% \\
Karaoke box & 0.1\% \\
Wangan Midnight & 0.1\% \\
Caporales & 0.0\% \\
Military band & 0.0\% \\
    \bottomrule
\end{tabular}

%% file: tables/new_domains.tex
\begin{tabular}{ll}
\toprule
Domain & Example Categories (\auc) \\
\midrule

Vehicles &  {\scriptsize  \begin{tabular}{@{}l@{}} Crossover SUV (70); Sedan (69);\\ Minivan (69); Station wagon (68) \end{tabular} } \\
Walkthroughs & {\scriptsize  \begin{tabular}{@{}l@{}}
Hotel Suite (71); Apartment (69); \\ Dining room (68); Living room (68)
\end{tabular} } \\
Advertising & 
{\scriptsize  \begin{tabular}{@{}l@{}}
Advertising (68); Television advertisement (66); \\ Infomercial (63)
\end{tabular} }
  \\
Tech Reviews &
{\scriptsize  \begin{tabular}{@{}l@{}}
CNET (66), Netbook (65), Asus (63) \\
IPhone 5S (64), MacBook (64)
\end{tabular} }
\\
Toys &
{\scriptsize  \begin{tabular}{@{}l@{}}
Funko (66); Monster High (64); Figurine (64) \\
Action Figure (64)

\end{tabular} }
\\
Appliances & 
{\scriptsize  \begin{tabular}{@{}l@{}}
Home appliance (65); Washing machine (64);\\ Kitchen stove (64)
\end{tabular} } \\
Places & 
{\scriptsize  \begin{tabular}{@{}l@{}}
Greenhouse (62)
University (61);\\ Amusement park (59) 
\end{tabular} } \\
\bottomrule
\end{tabular}

%% file: tables/crosstask_results.tex
\begin{table*}[t]
\resizebox{\textwidth}{!}{
\begin{tabular}{lc@{~~~~}c@{~~}c@{~~}c@{~~}c@{~~}c@{~~}c@{~~}c@{~~}c@{~~}c@{~~}c@{~~}c@{~~}c@{~~}c@{~~}c@{~~}c@{~~}c@{~~}c@{~~}|c} \toprule
    & \rotatebox{90}{\small Make} \rotatebox{90}{\small Kimchi Rice}  
    & \rotatebox{90}{\small Pickle} \rotatebox{90}{\small Cucumber}  
    & \rotatebox{90}{\small Make Banana} \rotatebox{90}{\small Ice Cream}  
    & \rotatebox{90}{\small Grill} \rotatebox{90}{\small Steak}  
    & \rotatebox{90}{\small Jack Up } \rotatebox{90}{\small Car}  
    & \rotatebox{90}{\small Make } \rotatebox{90}{\small Jello Shots}  
    & \rotatebox{90}{\small Change } \rotatebox{90}{\small Tire}  
    & \rotatebox{90}{\small Make } \rotatebox{90}{\small Lemonade}  
    & \rotatebox{90}{\small Add Oil } \rotatebox{90}{\small to Car}  
    & \rotatebox{90}{\small Make } \rotatebox{90}{\small Latte}  
    & \rotatebox{90}{\small Build } \rotatebox{90}{\small Shelves}  
    & \rotatebox{90}{\small Make } \rotatebox{90}{\small Taco Salad}  
    & \rotatebox{90}{\small Make } \rotatebox{90}{\small French Toast}  
    & \rotatebox{90}{\small Make } \rotatebox{90}{\small Irish Coffee}  
    & \rotatebox{90}{\small Make } \rotatebox{90}{\small Strawberry Cake}  
    & \rotatebox{90}{\small Make } \rotatebox{90}{\small Pancakes}  
    & \rotatebox{90}{\small Make } \rotatebox{90}{\small Meringue}  
    & \rotatebox{90}{\small Make } \rotatebox{90}{\small Fish Curry}  
    & \rotatebox{90}{ \textbf{Average} }
\\ \toprule
\newcite{zhukov2019cross} & 15.6 & 10.6 & 7.5 & 14.2 & 9.3 & 11.8 & 17.3 & 13.1 & 6.4 & 12.9 & 27.2 & 9.2 & 15.7 & 8.6 & 16.3 & 13.0 & 23.2 & 7.4 & 13.3 \\
Supervised upper-bound \cite{zhukov2019cross} & 19.1 & 25.3 & 38.0 & 37.5 & 25.7 & 28.2 & 54.3 & 25.8 & 18.3 & 31.2 & 47.7 & 12.0 & 39.5 & 23.4 & 30.9 & 41.1 & 53.4 & 17.3 & 31.6 \\
HowTo100M (1.2M videos) $\rightarrow$ Crosstask \cite{miech2019howto100m} & 33.5 & 27.1 & 36.6 & 37.9 & 24.1 & 35.6 & 32.7 & 35.1 & 30.7 & 28.5 & 43.2 & 19.8 & 34.7 & 33.6 & 40.4 & 41.6 & 41.9 & 27.4 & 33.6 \\
\rotatebox[origin=c]{180}{$\Lsh$} only 600K instructional videos & & & & & & & & & & & & & & & & & & & 32.6 \\
\rotatebox[origin=c]{180}{$\Lsh$} only 200K instructional videos & & & & & & & & & & & & & & & & & & & 31.1 \\
\midrule
Our HowTo100M (1.06M videos) $\rightarrow$ Crosstask & 24.5 & 30.0 & 39.9 & 32.0 & 27.0 & 37.2 & 33.6 & 33.5 & 24.4 & 27.7 & 44.7 & 19.1 & 32.9 & 31.7 & 35.3 & 46.6 & 43.4 & 22.9 & 32.6 \\
Our \yttrain (639K videos; 166K instr) $\rightarrow$ Crosstask & 21.5 & 24.7 & 35.2 & 26.2 & 19.6 & 29.5 & 25.8 & 30.1 & 20.9 & 22.9 & 32.7 & 18.4 & 26.7 & 27.2 & 31.0 & 43.3 & 37.7 & 20.5 & 27.4 \\
\bottomrule
\end{tabular}
}
\caption{Comparison between our simplified model and \newcite{miech2019howto100m}'s model on CrossTask, and the effect of pretraining the model on \yttrain instead of HowTo100M. Note that HowTo100M $\rightarrow$ Crosstask results are pre-trained on less data when compared to the original works due to video deletion.}
\label{tab:crosstask_replication}
\end{table*}

%% file: tables/supplementary_tables/10_categories.tex
\begin{tabular}{ll}
\toprule
Domain & Example Categories (\auc) \\
\midrule

Vehicles &  {\scriptsize  \begin{tabular}{@{}l@{}} Crossover SUV (73); Sedan (72);\\ Minivan (72); Station wagon (71) \end{tabular} } \\
Walkthroughs & {\scriptsize  \begin{tabular}{@{}l@{}}
Hotel Suite (74); Apartment (72); \\ Dining room (71); Living room (70)
\end{tabular} } \\
Advertising & 
{\scriptsize  \begin{tabular}{@{}l@{}}
Advertising (69); Television advertisement (67); \\ Infomercial (64)
\end{tabular} }
  \\
Tech Reviews &
{\scriptsize  \begin{tabular}{@{}l@{}}
CNET (67), Netbook (68), Asus (66) \\
IPhone 5S (66), MacBook (66)
\end{tabular} }
\\
Toys &
{\scriptsize  \begin{tabular}{@{}l@{}}
Funko (68); Monster High (66); Figurine (66) \\
Action Figure (66)

\end{tabular} }
\\
Appliances & 
{\scriptsize  \begin{tabular}{@{}l@{}}
Home appliance (67); Washing machine (66);\\ Kitchen stove (66)
\end{tabular} } \\
Places & 
{\scriptsize  \begin{tabular}{@{}l@{}}
Greenhouse (63)
University (63);\\ Amusement park (60) 
\end{tabular} } \\
\bottomrule
\end{tabular}

%% file: tables/supplementary_tables/30_categories.tex
\begin{tabular}{ll}
\toprule
Domain & Example Categories (\auc) \\
\midrule

Vehicles &  {\scriptsize  \begin{tabular}{@{}l@{}} Crossover SUV (72); Sedan (72);\\ Minivan (71); Station wagon (71) \end{tabular} } \\
Walkthroughs & {\scriptsize  \begin{tabular}{@{}l@{}}
Hotel Suite (74); Apartment (71); \\ Dining room (72); Living room (71)
\end{tabular} } \\
Advertising & 
{\scriptsize  \begin{tabular}{@{}l@{}}
Advertising (67); Television advertisement (65); \\ Infomercial (61)
\end{tabular} }
  \\
Tech Reviews &
{\scriptsize  \begin{tabular}{@{}l@{}}
CNET (65), Netbook (67), Asus (65) \\
IPhone 5S (64), MacBook (65)
\end{tabular} }
\\
Toys &
{\scriptsize  \begin{tabular}{@{}l@{}}
Funko (66); Monster High (65); Figurine (65) \\
Action Figure (66)

\end{tabular} }
\\
Appliances & 
{\scriptsize  \begin{tabular}{@{}l@{}}
Home appliance (65); Washing machine (64);\\ Kitchen stove (64)
\end{tabular} } \\
Places & 
{\scriptsize  \begin{tabular}{@{}l@{}}
Greenhouse (61)
University (60);\\ Amusement park (59) 
\end{tabular} } \\
\bottomrule
\end{tabular}

%% file: paper.bbl
\begin{thebibliography}{41}
\expandafter\ifx\csname natexlab\endcsname\relax\def\natexlab#1{#1}\fi

\bibitem[{Abu-El-Haija et~al.(2016)Abu-El-Haija, Kothari, Lee, Natsev,
  Toderici, Varadarajan, and Vijayanarasimhan}]{abu2016youtube}
Sami Abu-El-Haija, Nisarg Kothari, Joonseok Lee, Paul Natsev, George Toderici,
  Balakrishnan Varadarajan, and Sudheendra Vijayanarasimhan. 2016.
\newblock Youtube-8m: A large-scale video classification benchmark.
\newblock \emph{arXiv preprint arXiv:1609.08675}.

\bibitem[{Alayrac et~al.(2016)Alayrac, Bojanowski, Agrawal, Sivic, Laptev, and
  Lacoste-Julien}]{alayrac2016unsupervised}
Jean-Baptiste Alayrac, Piotr Bojanowski, Nishant Agrawal, Josef Sivic, Ivan
  Laptev, and Simon Lacoste-Julien. 2016.
\newblock Unsupervised learning from narrated instruction videos.
\newblock In \emph{CVPR}.

\bibitem[{Alayrac et~al.(2017)Alayrac, Laptev, Sivic, and
  Lacoste-Julien}]{alayrac2017joint}
Jean-Baptiste Alayrac, Ivan Laptev, Josef Sivic, and Simon Lacoste-Julien.
  2017.
\newblock Joint discovery of object states and manipulation actions.
\newblock In \emph{ICCV}.

\bibitem[{Amrani et~al.(2020)Amrani, Ben-Ari, Rotman, and
  Bronstein}]{amrani2020noise}
Elad Amrani, Rami Ben-Ari, Daniel Rotman, and Alex Bronstein. 2020.
\newblock Noise estimation using density estimation for self-supervised
  multimodal learning.
\newblock \emph{arXiv preprint arXiv:2003.03186}.

\bibitem[{Berg et~al.(2010)Berg, Berg, and Shih}]{berg2010automatic}
Tamara~L Berg, Alexander~C Berg, and Jonathan Shih. 2010.
\newblock Automatic attribute discovery and characterization from noisy web
  data.
\newblock In \emph{ECCV}.

\bibitem[{Bregler(1997)}]{bregler1997learning}
Christoph Bregler. 1997.
\newblock Learning and recognizing human dynamics in video sequences.
\newblock In \emph{Computer Society Conference on Computer Vision and Pattern
  Recognition}.

\bibitem[{Chang et~al.(2019)Chang, Huang, Sui, Fei-Fei, and
  Niebles}]{chang2019d3tw}
Chien-Yi Chang, De-An Huang, Yanan Sui, Li~Fei-Fei, and Juan~Carlos Niebles.
  2019.
\newblock D3tw: Discriminative differentiable dynamic time warping for weakly
  supervised action alignment and segmentation.
\newblock In \emph{CVPR}.

\bibitem[{Fouhey et~al.(2018)Fouhey, Kuo, Efros, and
  Malik}]{fouhey2018lifestyle}
David~F Fouhey, Wei-cheng Kuo, Alexei~A Efros, and Jitendra Malik. 2018.
\newblock From lifestyle vlogs to everyday interactions.
\newblock In \emph{CVPR}.

\bibitem[{Gu et~al.(2018)Gu, Sun, Ross, Vondrick, Pantofaru, Li,
  Vijayanarasimhan, Toderici, Ricco, Sukthankar, Schmid, and Malik}]{gu2018ava}
Chunhui Gu, Chen Sun, David~A Ross, Carl Vondrick, Caroline Pantofaru, Yeqing
  Li, Sudheendra Vijayanarasimhan, George Toderici, Susanna Ricco, Rahul
  Sukthankar, Cordelia Schmid, and Jitendra Malik. 2018.
\newblock {AVA}: A video dataset of spatio-temporally localized atomic visual
  actions.
\newblock In \emph{CVPR}.

\bibitem[{Gupta et~al.(2017)Gupta, Miao, Neves, and Metze}]{gupta2017visual}
Abhinav Gupta, Yajie Miao, Leonardo Neves, and Florian Metze. 2017.
\newblock Visual features for context-aware speech recognition.
\newblock In \emph{ICASSP}.

\bibitem[{Hessel et~al.(2018)Hessel, Mimno, and Lee}]{hessel2018quantifying}
Jack Hessel, David Mimno, and Lillian Lee. 2018.
\newblock Quantifying the visual concreteness of words and topics in multimodal
  datasets.
\newblock In \emph{NAACL}.

\bibitem[{Hill and Korhonen(2014)}]{hill2014learning}
Felix Hill and Anna Korhonen. 2014.
\newblock Learning abstract concept embeddings from multi-modal data: Since you
  probably can’t see what {I} mean.
\newblock In \emph{EMNLP}.

\bibitem[{Huang* et~al.(2018)Huang*, Buch*, Dery, Garg, Fei-Fei, and
  Niebles}]{huang-buch-2018-finding-it}
De-An Huang*, Shyamal Buch*, Lucio Dery, Animesh Garg, Li~Fei-Fei, and
  Juan~Carlos Niebles. 2018.
\newblock Finding ``it'': Weakly-supervised, reference-aware visual grounding
  in instructional videos.
\newblock In \emph{CVPR}.

\bibitem[{Huang et~al.(2017)Huang, Lim, Fei-Fei, and
  Carlos~Niebles}]{huang2017unsupervised}
De-An Huang, Joseph~J Lim, Li~Fei-Fei, and Juan Carlos~Niebles. 2017.
\newblock Unsupervised visual-linguistic reference resolution in instructional
  videos.
\newblock In \emph{CVPR}.

\bibitem[{Ignat et~al.(2019)Ignat, Burdick, Deng, and
  Mihalcea}]{ignat2019identifying}
Oana Ignat, Laura Burdick, Jia Deng, and Rada Mihalcea. 2019.
\newblock Identifying visible actions in lifestyle vlogs.
\newblock In \emph{ACL}.

\bibitem[{Kay et~al.(2017)Kay, Carreira, Simonyan, Zhang, Hillier,
  Vijayanarasimhan, Viola, Green, Back, Natsev, Suleyman, and
  Zisserman}]{kay2017kinetics}
Will Kay, Jo\~ao Carreira, Karen Simonyan, Brian Zhang, Chloe Hillier,
  Sudheendra Vijayanarasimhan, Fabio Viola, Tim Green, Trevor Back, Paul
  Natsev, Mustafa Suleyman, and Andrew Zisserman. 2017.
\newblock The kinetics human action video dataset.
\newblock \emph{arXiv preprint arXiv:1705.06950}.

\bibitem[{Kingma and Ba(2015)}]{kingma2014adam}
Diederik~P Kingma and Jimmy Ba. 2015.
\newblock Adam: A method for stochastic optimization.
\newblock In \emph{ICLR}.

\bibitem[{Kuehne et~al.(2019)Kuehne, Iqbal, Richard, and
  Gall}]{kuehne2019mining}
Hilde Kuehne, Ahsan Iqbal, Alexander Richard, and Juergen Gall. 2019.
\newblock Mining youtube-a dataset for learning fine-grained action concepts
  from webly supervised video data.
\newblock \emph{arXiv preprint arXiv:1906.01012}.

\bibitem[{Lu et~al.(2008)Lu, Zhang, Tian, and Ma}]{lu2008high}
Yijuan Lu, Lei Zhang, Qi~Tian, and Wei-Ying Ma. 2008.
\newblock What are the high-level concepts with small semantic gaps?
\newblock In \emph{CVPR}.

\bibitem[{Malmaud et~al.(2015)Malmaud, Huang, Rathod, Johnston, Rabinovich, and
  Murphy}]{malmaud2015s}
Jonathan Malmaud, Jonathan Huang, Vivek Rathod, Nick Johnston, Andrew
  Rabinovich, and Kevin Murphy. 2015.
\newblock What's cookin'? interpreting cooking videos using text, speech and
  vision.
\newblock In \emph{NAACL}.

\bibitem[{Menon and Elkan(2011)}]{menon2011link}
Aditya~Krishna Menon and Charles Elkan. 2011.
\newblock Link prediction via matrix factorization.
\newblock In \emph{ECML-PKDD}.

\bibitem[{Miech et~al.(2020)Miech, Alayrac, Smaira, Laptev, Sivic, and
  Zisserman}]{miech2019end}
Antoine Miech, Jean-Baptiste Alayrac, Lucas Smaira, Ivan Laptev, Josef Sivic,
  and Andrew Zisserman. 2020.
\newblock End-to-end learning of visual representations from uncurated
  instructional videos.
\newblock In \emph{CVPR}.

\bibitem[{Miech et~al.(2019)Miech, Zhukov, Alayrac, Tapaswi, Laptev, and
  Sivic}]{miech2019howto100m}
Antoine Miech, Dimitri Zhukov, Jean-Baptiste Alayrac, Makarand Tapaswi, Ivan
  Laptev, and Josef Sivic. 2019.
\newblock How{T}o100{M}: {L}earning a {T}ext-{V}ideo {E}mbedding by {W}atching
  {H}undred {M}illion {N}arrated {V}ideo {C}lips.
\newblock In \emph{ICCV}.

\bibitem[{Moriya et~al.(2019)Moriya, Sanabria, Metze, and
  Jones}]{moriya2019grounding}
Yasufumi Moriya, Ramon Sanabria, Florian Metze, and Gareth~JF Jones. 2019.
\newblock Grounding object detections with transcriptions.
\newblock \emph{arXiv preprint arXiv:1906.06147}.

\bibitem[{Parikh and Grauman(2011)}]{parikh2011interactively}
Devi Parikh and Kristen Grauman. 2011.
\newblock Interactively building a discriminative vocabulary of nameable
  attributes.
\newblock In \emph{CVPR}.

\bibitem[{Rendle et~al.(2009)Rendle, Freudenthaler, Gantner, and
  Schmidt-Thieme}]{rendle2012bpr}
Steffen Rendle, Christoph Freudenthaler, Zeno Gantner, and Lars Schmidt-Thieme.
  2009.
\newblock Bpr: Bayesian personalized ranking from implicit feedback.
\newblock In \emph{UAI}.

\bibitem[{Roberts et~al.(2020)Roberts, Raffel, and Shazeer}]{roberts2020much}
Adam Roberts, Colin Raffel, and Noam Shazeer. 2020.
\newblock How much knowledge can you pack into the parameters of a language
  model?
\newblock \emph{arXiv preprint arXiv:2002.08910}.

\bibitem[{Sanabria et~al.(2018)Sanabria, Caglayan, Palaskar, Elliott, Barrault,
  Specia, and Metze}]{sanabria2018how2}
Ramon Sanabria, Ozan Caglayan, Shruti Palaskar, Desmond Elliott, Lo{\"\i}c
  Barrault, Lucia Specia, and Florian Metze. 2018.
\newblock How2: a large-scale dataset for multimodal language understanding.
\newblock In \emph{NeurIPS Workshops}.

\bibitem[{Sener et~al.(2015)Sener, Zamir, Savarese, and
  Saxena}]{sener2015unsupervised}
Ozan Sener, Amir~R Zamir, Silvio Savarese, and Ashutosh Saxena. 2015.
\newblock Unsupervised semantic parsing of video collections.
\newblock In \emph{ICCV}.

\bibitem[{Sun et~al.(2019{\natexlab{a}})Sun, Baradel, Murphy, and
  Schmid}]{sun2019contrastive}
Chen Sun, Fabien Baradel, Kevin Murphy, and Cordelia Schmid.
  2019{\natexlab{a}}.
\newblock Contrastive bidirectional transformer for temporal representation
  learning.
\newblock \emph{arXiv preprint arXiv:1906.05743}.

\bibitem[{Sun et~al.(2019{\natexlab{b}})Sun, Myers, Vondrick, Murphy, and
  Schmid}]{sun2019videobert}
Chen Sun, Austin Myers, Carl Vondrick, Kevin Murphy, and Cordelia Schmid.
  2019{\natexlab{b}}.
\newblock {VideoBERT}: A joint model for video and language representation
  learning.
\newblock In \emph{ICCV}.

\bibitem[{Sun et~al.(2017)Sun, Shrivastava, Singh, and
  Gupta}]{sun2017revisiting}
Chen Sun, Abhinav Shrivastava, Saurabh Singh, and Abhinav Gupta. 2017.
\newblock Revisiting unreasonable effectiveness of data in deep learning era.
\newblock In \emph{ICCV}.

\bibitem[{Szegedy et~al.(2015)Szegedy, Liu, Jia, Sermanet, Reed, Anguelov,
  Erhan, Vanhoucke, and Rabinovich}]{inception_v1}
Christian Szegedy, Wei Liu, Yangqing Jia, Pierre Sermanet, Scott~E. Reed,
  Dragomir Anguelov, Dumitru Erhan, Vincent Vanhoucke, and Andrew Rabinovich.
  2015.
\newblock Going deeper with convolutions.
\newblock In \emph{CVPR}.

\bibitem[{Tang et~al.(2019)Tang, Ding, Rao, Zheng, Zhang, Zhao, Lu, and
  Zhou}]{tang2019coin}
Yansong Tang, Dajun Ding, Yongming Rao, Yu~Zheng, Danyang Zhang, Lili Zhao,
  Jiwen Lu, and Jie Zhou. 2019.
\newblock {COIN}: A large-scale dataset for comprehensive instructional video
  analysis.
\newblock In \emph{CVPR}.

\bibitem[{Wang et~al.(2019)Wang, Wang, Chen, and Jin}]{wang2019youmakeup}
Weiying Wang, Yongcheng Wang, Shizhe Chen, and Qin Jin. 2019.
\newblock Youmakeup: A large-scale domain-specific multimodal dataset for
  fine-grained semantic comprehension.
\newblock In \emph{EMNLP}.

\bibitem[{Xie et~al.(2018)Xie, Sun, Huang, Tu, and Murphy}]{xie2018rethinking}
Saining Xie, Chen Sun, Jonathan Huang, Zhuowen Tu, and Kevin Murphy. 2018.
\newblock Rethinking spatiotemporal feature learning: Speed-accuracy trade-offs
  in video classification.
\newblock In \emph{ECCV}.

\bibitem[{Yanai and Barnard(2005)}]{yanai2005image}
Keiji Yanai and Kobus Barnard. 2005.
\newblock Image region entropy: a measure of visualness of web images
  associated with one concept.
\newblock In \emph{ACM MM}.

\bibitem[{Yu et~al.(2014)Yu, Jiang, and Hauptmann}]{yu2014instructional}
Shoou-I Yu, Lu~Jiang, and Alexander Hauptmann. 2014.
\newblock Instructional videos for unsupervised harvesting and learning of
  action examples.
\newblock In \emph{ACM MM}.

\bibitem[{Zhu and Yang(2020)}]{zhuactbert}
Linchao Zhu and Yi~Yang. 2020.
\newblock Actbert: Learning global-local video-text representations.
\newblock In \emph{CVPR}.

\bibitem[{Zhu et~al.(2015)Zhu, Kiros, Zemel, Salakhutdinov, Urtasun, Torralba,
  and Fidler}]{zhu2015aligning}
Yukun Zhu, Ryan Kiros, Rich Zemel, Ruslan Salakhutdinov, Raquel Urtasun,
  Antonio Torralba, and Sanja Fidler. 2015.
\newblock Aligning books and movies: Towards story-like visual explanations by
  watching movies and reading books.
\newblock In \emph{ICCV}.

\bibitem[{Zhukov et~al.(2019)Zhukov, Alayrac, Cinbis, Fouhey, Laptev, and
  Sivic.}]{zhukov2019cross}
Dimitri Zhukov, Jean-Baptiste Alayrac, Ramazan~Gokberk Cinbis, David Fouhey,
  Ivan Laptev, and Josef Sivic. 2019.
\newblock Cross-task weakly supervised learning from instructional videos.
\newblock In \emph{CVPR}.

\end{thebibliography}


\begin{thebibliography}{7}
\expandafter\ifx\csname natexlab\endcsname\relax\def\natexlab#1{#1}\fi

\bibitem[{Kay et~al.(2017)Kay, Carreira, Simonyan, Zhang, Hillier,
  Vijayanarasimhan, Viola, Green, Back, Natsev, Suleyman, and
  Zisserman}]{kay2017kinetics}
Will Kay, Jo\~ao Carreira, Karen Simonyan, Brian Zhang, Chloe Hillier,
  Sudheendra Vijayanarasimhan, Fabio Viola, Tim Green, Trevor Back, Paul
  Natsev, Mustafa Suleyman, and Andrew Zisserman. 2017.
\newblock The kinetics human action video dataset.
\newblock \emph{arXiv preprint arXiv:1705.06950}.

\bibitem[{Miech et~al.(2019)Miech, Zhukov, Alayrac, Tapaswi, Laptev, and
  Sivic}]{miech2019howto100m}
Antoine Miech, Dimitri Zhukov, Jean-Baptiste Alayrac, Makarand Tapaswi, Ivan
  Laptev, and Josef Sivic. 2019.
\newblock How{T}o100{M}: {L}earning a {T}ext-{V}ideo {E}mbedding by {W}atching
  {H}undred {M}illion {N}arrated {V}ideo {C}lips.
\newblock In \emph{ICCV}.

\bibitem[{Sun et~al.(2019)Sun, Myers, Vondrick, Murphy, and
  Schmid}]{sun2019videobert}
Chen Sun, Austin Myers, Carl Vondrick, Kevin Murphy, and Cordelia Schmid. 2019.
\newblock {VideoBERT}: A joint model for video and language representation
  learning.
\newblock In \emph{ICCV}.

\bibitem[{Sun et~al.(2017)Sun, Shrivastava, Singh, and
  Gupta}]{sun2017revisiting}
Chen Sun, Abhinav Shrivastava, Saurabh Singh, and Abhinav Gupta. 2017.
\newblock Revisiting unreasonable effectiveness of data in deep learning era.
\newblock In \emph{ICCV}.

\bibitem[{Szegedy et~al.(2015)Szegedy, Liu, Jia, Sermanet, Reed, Anguelov,
  Erhan, Vanhoucke, and Rabinovich}]{inception_v1}
Christian Szegedy, Wei Liu, Yangqing Jia, Pierre Sermanet, Scott~E. Reed,
  Dragomir Anguelov, Dumitru Erhan, Vincent Vanhoucke, and Andrew Rabinovich.
  2015.
\newblock Going deeper with convolutions.
\newblock In \emph{CVPR}.

\bibitem[{Xie et~al.(2018)Xie, Sun, Huang, Tu, and Murphy}]{xie2018rethinking}
Saining Xie, Chen Sun, Jonathan Huang, Zhuowen Tu, and Kevin Murphy. 2018.
\newblock Rethinking spatiotemporal feature learning: Speed-accuracy trade-offs
  in video classification.
\newblock In \emph{ECCV}.

\bibitem[{Zhukov et~al.(2019)Zhukov, Alayrac, Cinbis, Fouhey, Laptev, and
  Sivic.}]{zhukov2019cross}
Dimitri Zhukov, Jean-Baptiste Alayrac, Ramazan~Gokberk Cinbis, David Fouhey,
  Ivan Laptev, and Josef Sivic. 2019.
\newblock Cross-task weakly supervised learning from instructional videos.
\newblock In \emph{CVPR}.

\end{thebibliography}
